\documentclass[sigconf]{acmart}
\renewcommand\footnotetextcopyrightpermission[1]{} 
\def\ie{{\em i.e.}}
\def\eg{{\em e.g.}}
\def\etal{{\em et al.}}

\newcommand{\figref}[1]{Fig. \ref{#1}}
\newcommand{\tabref}[1]{Tab. \ref{#1}}

\newcommand{\mc}[1]{\mathcal{#1}}

\newcommand{\br}[1]{\bm{\mathrm{#1}}}
\graphicspath{{./}}

\usepackage{epstopdf}

\usepackage{ragged2e}
\usepackage{booktabs}
\usepackage{color}
\usepackage{multirow}
\usepackage{bm}
\usepackage{bbm}
\usepackage[misc]{ifsym}

\usepackage{multirow}
\usepackage{float}
\usepackage{stfloats}

\usepackage{amssymb}

\usepackage{pifont}
\newcommand{\cmark}{\ding{51}}%
\newcommand{\xmark}{\ding{55}}%

\AtBeginDocument{%
  \providecommand\BibTeX{{%
    \normalfont B\kern-0.5em{\scshape i\kern-0.25em b}\kern-0.8em\TeX}}}

\setcopyright{acmcopyright}
\copyrightyear{2020}
\acmYear{2020}
\setcopyright{acmlicensed}
\acmConference[MM '20]{Proceedings of ACM Multimedia conference	(MM20)}{12 - 16 October 2020}{Seattle, United States}
\acmBooktitle{Proceedings of ACM Multimedia conference (MM'20)}
\acmPrice{15.00}
\acmDOI{10.475/123_4}
\acmISBN{978-1-4503-9999-9/18/06}

\begin{document}

\title{Cartoon Face Recognition: A Benchmark Dataset}

\author{Yi Zheng$^1$, Yifan Zhao$^2$, Mengyuan Ren$^1$, He Yan$^1$, Xiangju Lu$^{1,*}$, Junhui Liu$^1$, Jia Li$^{2,*}$
}
\affiliation{
    \institution{$^1$iQIYI, Inc \\
	 }
	\institution{$^2$State Key Laboratory of Virtual Reality Technology and Systems, SCSE, Beihang University\\
	 }
}
\thanks{$^*$ Correspondence should be addressed to Xiangju Lu and Jia Li (E-mail: \textsuperscript{}luxiangju@qiyi.com, \textsuperscript{}jiali@buaa.edn.cn).}

%
\renewcommand{\shortauthors}{Yi Zheng and Yifan Zhao, et al.}

\begin{abstract}
Recent years have witnessed increasing attention in cartoon media, powered by the strong demands of industrial applications.
As the first step to understand this media, cartoon face recognition is a crucial but less-explored task with few datasets proposed. In this work, we first present a new challenging benchmark dataset, consisting of 389,678 images of 5,013 cartoon characters annotated with identity, bounding box, pose, and other auxiliary attributes. The dataset, named iCartoonFace, is currently the largest-scale, high-quality, rich-annotated, and spanning multiple occurrences in the field of image recognition, including near-duplications, occlusions, and appearance changes. In addition, we provide two types of annotations for cartoon media,~\ie, face recognition, and face detection, with the help of a semi-automatic labeling algorithm.
To further investigate this challenging dataset, we propose a multi-task domain adaptation approach that jointly utilizes the human and cartoon domain knowledge with three discriminative regularizations.
We hence perform a benchmark analysis of the proposed dataset and verify the superiority of the proposed approach in the cartoon face recognition task. We believe this public availability will attract more research attention in broad practical application scenarios.
\end{abstract}


\begin{teaserfigure}
\centering
	\includegraphics[width=1\columnwidth]{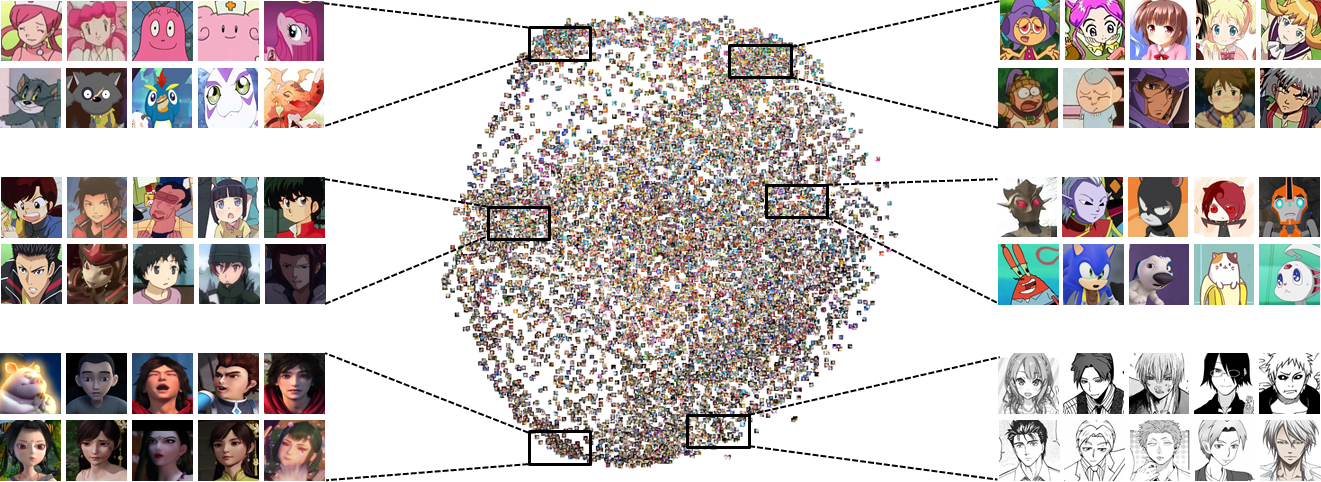}
	\caption{Illustration of iCartoonFace embedding. The proposed dataset consists of diverse data sources for face recognition and detection task. Dataset has been publicly available for promoting subsequent researches.}
	\label{fig:intro}
\end{teaserfigure}

\maketitle
\section{Introduction}
What helps one recognize a face? Despite the faces in real-world images, cartoon face is a vital part to understand and interact with the virtual world. Accurately recognizing these cartoon characters is an essential prerequisite for many vision applications, such as automatic editing, filming, advertisement recommendation, and computer-aided modeling~\cite{chen2002pictoon,elad2005simultaneous}. With the proposals of large datasets~\cite{huang2008labeled,yi2014learning}, deep models on human faces have achieved transpersonal accuracies, which greatly surpasses the conventional hand-crafted methods. For example, ArcFace~\cite{deng2019arcface} reached a precision of 99.83\% on LFW benchmark and the best accuracy on MegaFace~\cite{kemelmacher2016megaface} have also reached 99.39\%.
However, the performance gap is mainly achieved by the utilization of tremendous manual labeling datasets, which is extremely deficient in the cartoon media.

Over the years, cartoon media have shown its strong correlations with the real-world knowledge.
Artists create and imagine the cartoon characters based on real-world abstractions and thus the created faces share a lot of similarities with human faces. To answer the aforementioned question, two natural problems raise our concerns: 1) what is the desirable need in cartoon dataset? 2) what is the relationship between human faces and cartoon ones?

In this less-explored recognition of virtual media, few datasets~\cite{mishra2016iiit,huo2017webcaricature,fujimoto2016manga109,danbooru2018} have been proposed for specific purposes, which can be roughly grouped into two categories. The first category is the caricature dataset, which is fundamentally based on real human identities. These caricature images share strong similarities with the human portrait but exaggerate certain specific facial features. For example, WebCaricature~\cite{huo2017webcaricature} built a large photograph-caricature dataset consisting of 6,042 caricatures and 5,974 photographs from 252 persons. IIIT-CFW~\cite{mishra2016iiit} established a challenging dataset of 8,928 annotated unconstrained cartoon faces of 100 international celebrities. Both datasets share lots of similarities with the real-world human faces but with the variation of artistic styles. This would lead to a severe recognition problem, when regarding the same figure drawn by different artists as one class.
The second category focuses on the cartoon recognition task, with very few datasets proposed. Datasets of this category do not rely on real celebrities or actors, but follow the basic rules in constructing a face. In these virtual media and cartoon videos, most characters show exaggerated or humorous facial expressions, which brings us new challenges in recognizing the same identity.
For example, Manga109~\cite{fujimoto2016manga109} proposed a dataset for cartoon image retrieval and face detection, consisting of 21,142 images from 109 Japanese comic books.
Besides its lack of face recognition, this dataset mainly collected limited images and restricted in the Japanese style comics, which may not satisfy the demand of large-scale training data for deep learning approaches.
In addition, the cartoon dataset is required to contain substantial complex scenarios, thus is available and robust for the industrial application.
To concretely meet the first concern, a high-quality, representative, large-scale dataset for cartoon face is in high demand.

\begin{table*}[t]
    \begin{center}
     \caption{Summary of existing datasets related to cartoon recognition.}\label{tab:summary}
    \setlength{\tabcolsep}{1.0mm}
    \begin{tabular}{c|cccccc}
    \toprule
    Dataset&Type& \#images  & \#identities&Artistic style& Face Anno. & Atrribute Anno.\\
    \midrule
   Klare~\etal.~\cite{klare2012towards}&Caricature recognition&392&196&\xmark & \xmark &\xmark \\
   Abaci~\etal~\cite{abaci2015matching}&Caricature recognition&400&200&\xmark & \cmark &\xmark \\
   WebCaricature~\cite{huo2017webcaricature}& Caricature recognition&12,016&252&\xmark&\cmark& Facial Landmark \\
   IIIT-CFW~\cite{mishra2016iiit}&Caricature recognition&8,928&100&\xmark&\xmark&Pose (1D), Age \\
   Manga109~\cite{fujimoto2016manga109}&Cartoon detection\&retrival&21,142&-&Unified&\cmark&\xmark \\
   \textbf{iCartoonFace}&Cartoon recognition\&detection&\textbf{389,678}&\textbf{5,013}&Unified&\cmark&Pose (3D), Gender \\
    \bottomrule
    \end{tabular}
    \end{center}
\end{table*}

It is notable that even the most unrealistic cartoon faces are created with \textit{anthropomorphism}, indicating the correlation between the virtual media and realistic human images. Hence, for the second concern, we would like to examine how useful the human face can help the cartoon ones, including the recognition task and detection task. Based on this thought, we develop semi-automatic annotation procedures that make use of existing human faces as embedding knowledge, which serves as preliminary detectors and classifiers for the labeling process. On the other hand, the existing human faces could also help the recognition of cartoon faces, which serves as a teacher network and transfers the domain knowledge to the cartoon domains. Toward this end, a high-quality benchmark dataset and a learning approach for cartoon face recognition are needed to be proposed.

In this paper, we present iCartoonFace: a high-quality, large-scale, rich annotated benchmark dataset for cartoon face recognition. The iCartoonFace dataset consists of 389,678 images of 5,013 cartoon persons from public websites and online videos. In addition, we also provide 60,000 images of 109,810 faces with bounding boxes to form the detection dataset
The iCartoonFace dataset also provides auxiliary information, including 3d pose (yaw, pitch and roll angles), collection source, album, and gender (shown in~\tabref{tab:summary}). This dataset spans multiple occurrences in the cartoon recognition challenges, including near-duplication, inter-class diversity, illumination conditions, and appearance changes.
Besides this challenging benchmark, we thus set out to propose a multi-task domain adaptation approach to transfer the real-world faces to the cartoon media, which jointly regularizes the embedding space with three cues,~\ie, the classification constraint, unknown identity constraint and cross-domain constraint. Experimental evidences demonstrate the challenges in the face recognition benchmark and the improvement potential in utilizing real-world data.

Main contributions of this work can be summarized as:
1) We present a high-quality, large-scale, challenging benchmark dataset for cartoon face recognition and face detection, which consists of 389,678 images of 5,013 cartoon persons. We thus develop a semi-automatic assembling strategy to collect these diverse images from 1,302 albums.
2) We propose a multi-task domain adaptation framework to investigate the potential of transferring knowledge from human domains and cartoon domains.
3) We perform a benchmark analysis of state-of-the-arts and open-source for the subsequent researches. Further experiments demonstrate the effectiveness of the proposed method and the challenge of the benchmark.

\section{Related Work}

\subsection{Datasets}
Existing works~\cite{guo2016ms,huang2008labeled} pay many efforts in the real-world human faces. For example, Labeled Faces in the Wild (LFW)~\cite{huang2008labeled} database of face images is designed for studying the problem of unconstrained face recognition. The database contains more than 13,000 images of 5,749 characters. MegaFace~\cite{kemelmacher2016megaface} is a large scale face database of over 1 million faces photos from 690k persons. CASIA-WebFace~\cite{yi2014learning} contains more than 10k identities and about 500k images for unlimited face recognition. CAS-PEAL~\cite{caspeal} contains more than 30k images of 1,040 persons for constrained face recognition, mainly includes gestures, expressions, lighting variations.

There are also some datasets~\cite{klare2012towards,abaci2015matching} related to cartoon and caricature recognition.
For example, IIIT-CFW~\cite{mishra2016iiit} contains 8,928 annotated cartoon faces of 100 international celebrities. IIIT-CFW can be used for a wide spectrum of problems due to the fact that it contains detailed annotations such as type of cartoon, pose, expression, age group, and \textit{etc.} WebCaricature~\cite{huo2017webcaricature} is a large photograph-caricature dataset consisting of 6,042 caricatures and 5,974 photographs from 252 persons collected from the web. For each image in the dataset, 17 labeled facial landmarks are provided. All two cartoon datasets are created from caricature. Manga109~\cite{fujimoto2016manga109} is a dataset of a variety of 109 Japanese comic books and created for detection. Danbooru~\cite{danbooru2018} simply gathered a large-scale collection of over 970k images of 70k identities. However, there are two main problems existing in this dataset. First, the annotations are roughly collected noisy labels, which is not annotated by human annotators. Even the best performed model~\cite{hu2018squeeze} can only reach the 37.3\% accuracy. Second, the same identity in this dataset may be created by different artists.
Thus a high-quality, manual labeling dataset for cartoon face recognition is in high demand. Thus we defined the two different types of artistic style in~\tabref{tab:summary}: \textit{Unified} indicates that characters are created in the same style or by the same artist.

\subsection{Face Recognition}
Face recognition can be seen as a sub-problem of image classification. Numerous models~\cite{he2016deep,huang2017densely,hu2018squeeze,xie2017aggregated} are proposed for solving image classification problem and get impressive results. For instance, ResNet~\cite{he2016deep} presented a residual learning framework to ease the training of networks that are substantially deeper than those used previously. DenseNet~\cite{huang2017densely} connected each layer to every other layer in a feed-forward fashion. SEnet~\cite{hu2018squeeze} recalibrated channel-wise feature responses by explicitly modeling interdependencies between channels.

Despite these network architectures, advanced face recognition algorithms~\cite{wang2020mis,masi2016pose,wang2018cosface,liu2017sphereface,wang2017normface,zheng2018ring} have also been proposed. For example, Wang~\etal~\cite{wang2017normface} proposed a hypersphere embedding strategy to enhance the face verification. Taigman~\etal~\cite{taigman2014deepface} presented a framework for recognizing the faces in the wild. SphereFace~\cite{liu2017sphereface} presented a novel loss that enables convolutional neural networks (CNNs) to learn angularly discriminative features. CosFace~\cite{wang2018cosface} maximized the decision margin in the cosine space. Arcface~\cite{deng2019arcface} maximized the decision margin in the angular space. However, these algorithms only study this problem in the image and face recognition in the real world. For object detection, focal loss~\cite{lin2017focal} adopts a re-weighting scheme to address the class imbalance problem, which is also widely used in the face recognition task. Some work~\cite{zheng2018ring,wang2017deep} presented a improved a softmax loss to
regularize the features in a normalized space. In addition, Yang~\etal~\cite{yang2020fan} proposed to embed the landmark information in a deep network to help the recognition progress.

\subsection{Multi-task Learning}
Zhang~\etal~\cite{zhang2017survey} made a detailed survey  of general multitask learning (MTL). More specifically related to our work,~\cite{zhang2012convex} solved a convex optimization problem to estimate task relationships, while~\cite{pentina2017multi} analyzed the weighted sum loss algorithm and its applications in the online, active and transductive scenarios. Moreover,~\cite{kendall2018multi} proposed probabilistic models through the construction of a task covariance matrix or estimate the multitask likelihood from a deep Bayes model. In the face recognition field, HyperFace~\cite{ranjan2017hyperface} proposed to classify a given image region as face or non-face, estimate the head pose, locate face landmarks and recognize gender in one network. Different from these aforementioned approaches, in this paper, we designed to propose a multi-task domain adaptation framework to regularize the embedding vectors with reasonable decision borders.

\begin{figure*}[t]
	\centering
	\includegraphics[width=\textwidth]{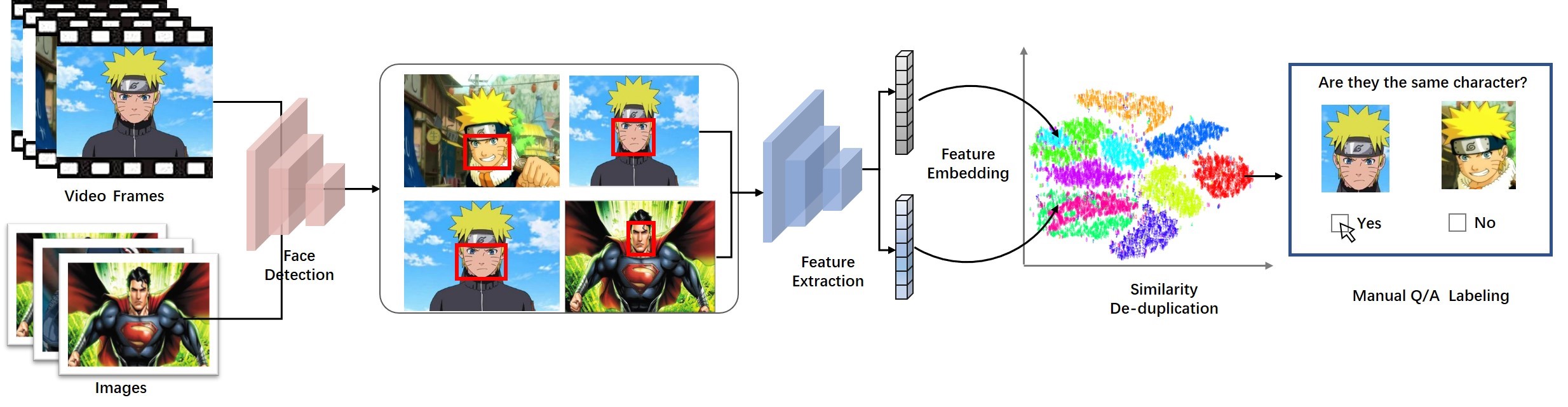}
	\caption{The overview of semi-automatic assembling process. Instead of labeling IDs, human annotators only need to answer the T/F questions.}
	\label{fig:flowchart}
\end{figure*}

\section{Dataset Construction and Analysis}

\subsection{Semi-automatic Assembling Process}
To reduce the labeling burden, we develop a semi-automatic
algorithm to collect and annotate the iCartoonFace dataset (See Fig. 3). Our framework can be conducted in the following three stages: 1) hierarchical data collection; 2) data filtering process; 3) Q/A manual annotation.

\textbf{Hierarchical Data Collection.} The iCartoonFace dataset is collected by hierarchical manners (from cartoon album names to cartoon person names, and finally to cartoon person images). We first form a cartoon album name list regarding the rank\footnote{https://en.wikipedia.org/}. Secondly, we obtain the main characters from the internet based on the album name list. Hence a list of cartoon persons and corresponding albums could be obtained. Thirdly, we download publicly available images from multimodal media, including public images, comic books, and video sources. In this manner, millions of images with noisy labels are obtained for subsequent data filtering process.

\begin{figure}[t]
	\includegraphics[width=1\linewidth]{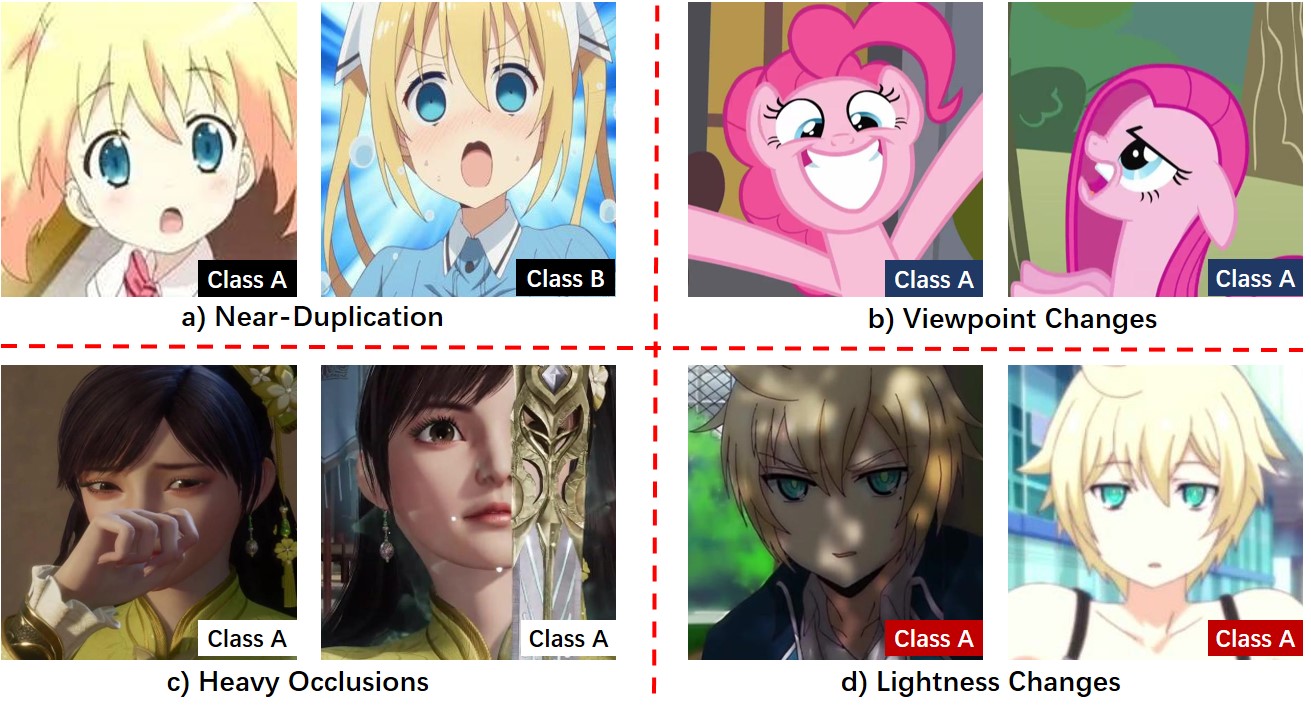}
	\caption{Four representative challenges in iCartoonFace: near-duplication, viewpoint changes, heavy occlusions, lightness changes.}
	\label{fig:challenge}
\end{figure}

\textbf{Guided Data Filtering.}
 In fact, there are tremendous irrelevant or duplicate data in downloaded images, which brings us a great challenge to select valid data, especially without any prior knowledge. Hence we resort to the manual labeled human faces, which provide coarse filtering for the useless samples. In other words, we resort to two existing human knowledge to help the data filtering process,~\ie, face detection filtering, and face recognition filtering.

 Hence, we build a data filtering process that consists of a face detection branch and a feature extraction branch. In the face detection part, we firstly manually labeled 60,000 images with 109,810 cartoon face bounding box as our cartoon face detection dataset, of which 80\% was used as the trainset and 20\% was used as the test set. We mix this part of data with the real human faces, resulting in a final 200,000 images with 500,000 bounding boxes to enhance the detection part.
 We adopt RetinaNet~\cite{lin2017focal} as our detection backbone and achieved 89\% mAP in 0.5 IoU on the test set. In the feature extraction part, we initially use Arcface model~\cite{deng2019arcface} pre-trained on existing human face recognition datasets as our feature extraction model. With the increase of cartoon datasets, the model can be jointly trained and the performances are improved steadily. The success of the filtering process in turn verifies the correlations of the real-world data and virtual ones.

\textbf{Q/A Manual Annotation.}
We developed a Q/A system to manually annotate the identity information of cartoon faces. In the annotation page (see ~\figref{fig:flowchart}), one part shows a reference image, and the other part displays the image to be labeled. The annotators needed to determine whether each new image shares the same identity with the reference image. The reference image is an identity picture provided by the expert based on the cartoon album name and cartoon person name to which the cartoon person belongs. In our dataset, 5,013 reference pictures are included, meaning that there is one probe for each identity.

\begin{figure*}[t]
	\centering
	\includegraphics[width=0.95\textwidth]{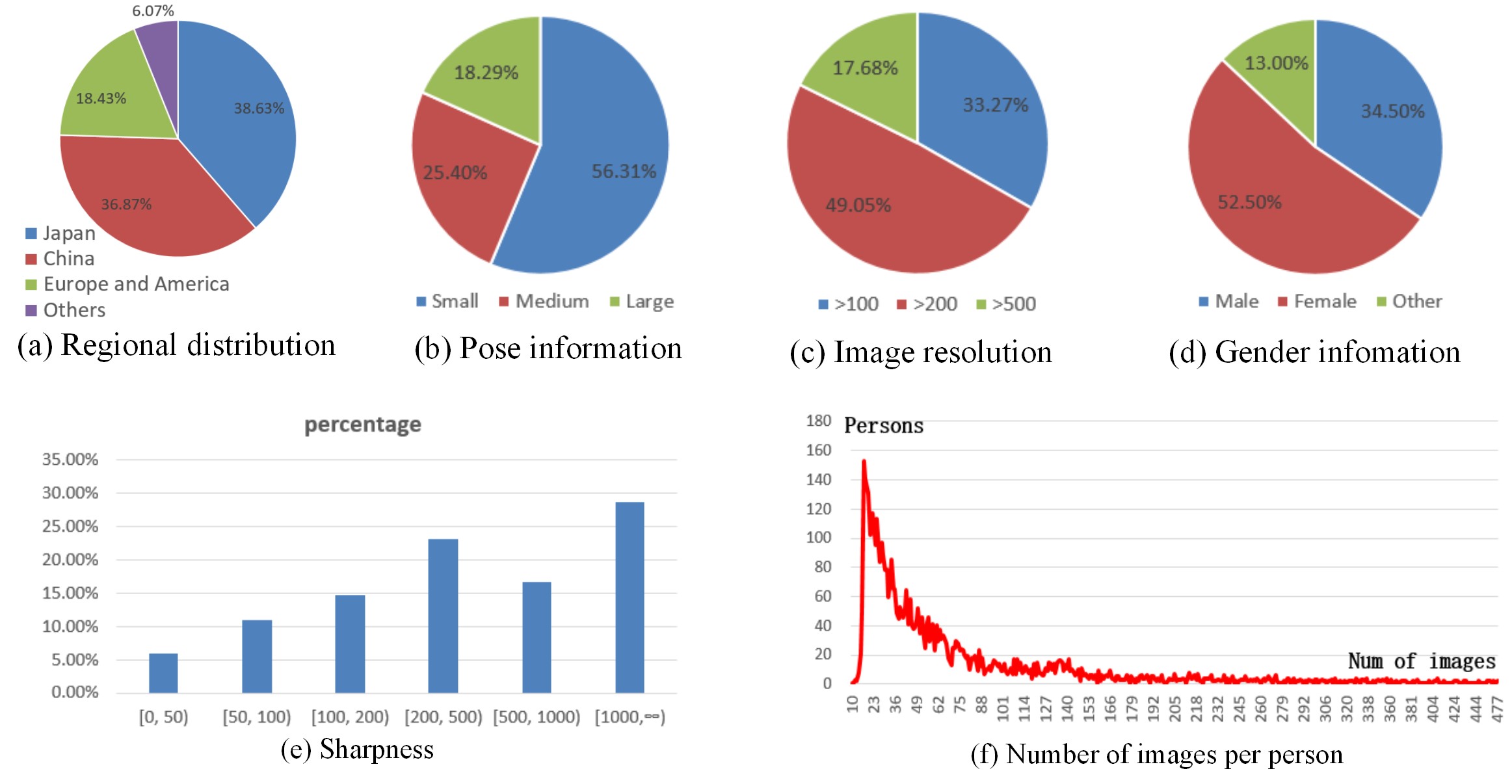}
	\caption{iCartoonFace dataset statistics. We present region distributions of identities, pose annotations, image resolution distribution, sharpness and number of samples per identity.}
	\label{fig:dataset}
\end{figure*}

\subsection{Dataset Statistics}
\textbf{Large scale.} The iCartoonFace dataset consists of 389,678 images of 5,013 cartoon persons coming from 1,302 cartoon albums. To our knowledge, this is currently the largest manual annotated image dataset for cartoon face recognition. As shown in Fig. 4 (a), the dataset cartoon persons are widely distributed in Japan, China, Europe and America.

\textbf{In the wild/long-tailed.} The dataset is created naturally, which the distribution is long-tailed in~\figref{fig:dataset} (f). 50\% of the identities owns less than 30 images, while some of the cases even owns about 500 images.

\textbf{High quality.} After the dataset was manually labeled, we perform a cross-checking method, and the re-checking error rate is guaranteed to be less than 5\%. \figref{fig:dataset} (c) shows that the resolution of images are more than $100 \times 100$ and more than 65\% of them are larger than $200 \times 200$. The sharpness of images is calculated by the Laplacian metric, and the values of most samples are more than 100 as shown in~\figref{fig:dataset} (e), to ensure the clearness and sharpness of image boundaries.

\textbf{Rich attributes.} It provides information such as the face bounding box, identity, region, pose, and gender in each image. The statistical information about pose and gender is shown in~\figref{fig:dataset} (b) and (d). Random 10,000 samples are selected and annotated with 3D pose information,~\ie. yaw, pitch and roll angles. 66\% samples have small angle less than 30 degrees, and about 8\% samples have large angle with more than 60 degrees up to 90 degrees.

\subsection{Challenges and Tasks}

\textbf{Challenges.}
We visualize four representative challenges of cartoon data in the wild (\figref{fig:challenge}):
a) \textit{Near-Duplication}: two images with different IDs are in a very similar appearance. This motivates the algorithms to be aware of subtle local differences.
b) \textit{Viewpoint Changes}: viewpoint changes of the same character brings us new challenges in recognizing one person.
c) \textit{Heavy Occlusions}: the faces might be occluded by other objects of the scenarios. Algorithms need to extract the most discriminative features for recognition.
d) \textit{Lightness Changes}: there are other kinds of variations, including the lightness and resolution changes. All these cues force the algorithms to be robustness for different scenarios.

\textbf{Face recognition.}
Following the pioneer human face dataset~\cite{kemelmacher2016megaface}, we proposed an identification task to benchmark the performance of the cartoon face recognition algorithms.
In the identification task setting, given a probe photo, and a gallery containing at least one image of the same cartoon person, the algorithm needs to rank orders all images in the gallery based on similarity to the probe.
Specifically, the probe set includes $N$ cartoon persons and each cartoon person has $M$ images. The algorithm then tests each of the $M$ images per cartoon person by adding it to the gallery of distractors and use each of the other $M-1$ photos as a probe. Results are presented with the identification rate of rank-K. For the identification test, a gallery set with 2,500 images with 2,500 different persons is created. To ensure a fair identification, the gallery set is created by the persons that their identity does not appear in the training set and probe set. The identification probe set contains 20,000 images from 2,000 persons (1,200 of person identity which in training set and 800 of person identity which not in training set). The number of samples for each person ranges from 5 to 17.

\textbf{Face detection.}
To construct a high-quality dataset, we select 50,000 images of 91,163 faces as training set and 10,000 images of 18,647 faces as testing set. We use mAP (mean Average Precision) as evaluation protocal,~\ie, $mAP=\frac{\sum_i^C AP_i}{C}$. Remarkably, this is also the largest cartoon face detection dataset to date.
All the datasets for these two tasks will be publicly available, and the details can be found in supplementary.

\begin{figure*}[t]
	\centering
	\includegraphics[width=1.0\textwidth]{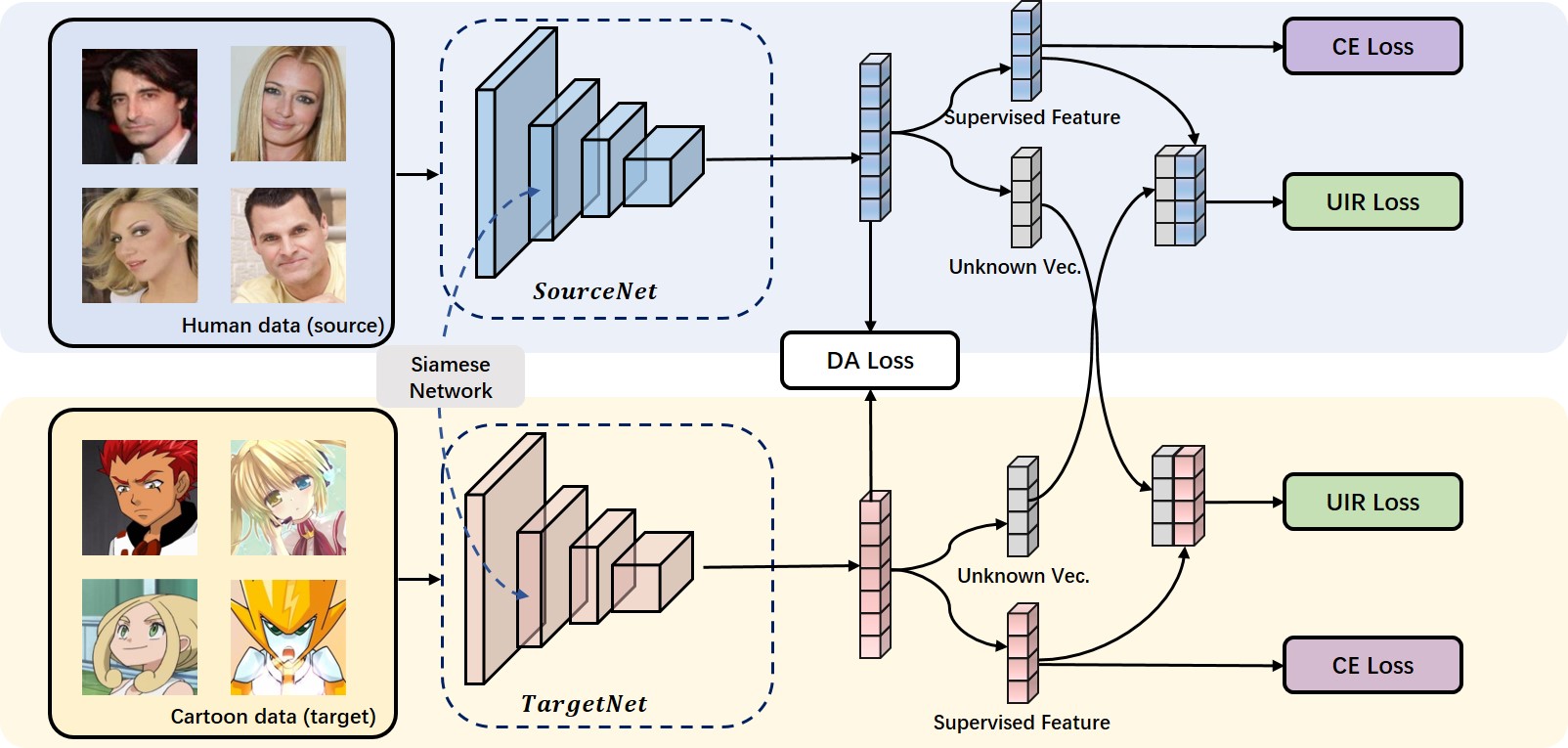}
	\caption{The overview of the proposed approach. Our framework is composed of three important constraints: the classifier of two domains, the unknown identity rejection and domain adaption adversarial loss.}
	\label{fig:framework}
\end{figure*}

\section{A Baseline Approach}

\subsection{Problem Formulation}

As mentioned above, the cartoon faces are derived from abstractions of human faces, thus share a lot of similar structures and common knowledge. In theory, our basic idea is to utilize the domain data from real-world human faces, helping to partition classification hyperplanes in cartoon domain.

To achieve this, we design three meaningful cues: 1) We designed a Siamese network to conduct the training procedure and optimize the classification loss of two domains per batch. As shown in~\figref{fig:framework}, these two networks are built in a weight-sharing manner and regularized with additional training data from the human domain. 2) As the labeled data is in the one-hot form, when re-partition the classification hyperplanes, the classification loss would force the predicted distribution to be a one-hot vector,~\eg,$[0,...,0,1]$. This optimization would lead to severe overfitting, especially when testing on unseen identities. To solve this issue, we adopt the unknown identity loss to smooth the predicted distributions with the help of data from other domains. That is to say, when smoothing the distribution of cartoon faces, we adopt human faces as unlabeled data for this optimization.
3) Inspired by the gradient reversal layer, we develop to use a domain classifier in an adversarial manner. A general classifier for both domains should not show specific domain features in the high-level layers,~\ie, whatever the input image is from, the distribution of final prediction should be similar. Our motif is to fool the classifier to do not be aware of its domain and to find a generalized classification hyper-planes.

Let $\mathbb{H}$ represents the human face domain and $\mathbb{C}$ be the cartoon face domain. We thus put the source training data $\mc{D}_h \sim \mathbb{H}$ and target data $\mc{D}_c \sim \mathbb{C}$ in one batch to conduct this training procedure.
Let $\br{T}=(\br{x},\br{y},\br{z})$ represents a typical triplet sampled from these two domains. In this triplet, $\br{x}$ denotes the input image, $\br{y}$ denotes the identity label of image $\br{x}$, and $\br{z}$ indicates whether the image belongs to a cartoon or a human face. When the input image is a cartoon face $\mathbb{C}$, $\br{z}$ is set as 1 and otherwise 0. With the given input, our aim is to predict a generalized distribution $\br{p}$ of both domains with the three aforementioned cues.

\subsection{Learning objective}
As shown in~\figref{fig:framework}, with an input image $\br{x}$ forward-propagated through the network, three discriminative classifiers can be obtained. Considering that these two networks are designed in a weight-sharing trend, in the testing phase, we do not introduce any additional computation cost.
With these classifiers, the total loss function $\mc{L}$ is composed of three parts: classification loss $\mc{L}_{cls}$, unknown identity rejection loss $\mc{L}_{uir}$, and domain transfer loss $\mc{L}_{da}$.

\textbf{Classification loss}. To obtain a discriminative feature extractor, we adopt the classification loss to regularize both cartoon and human face classifiers, represented as $\mc{L}^{c}_{cls}$ and $\mc{L}^{h}_{cls}$, respectively. For $\forall (\br{x}_c,\br{y}_c) \in \mc{D}_c, (\br{x}_h,\br{y}_h) \in \mc{D}_h$, a typical classification loss can be presented as:
\begin{equation}
\begin{split}
\mc{L}_{cls}&=\mc{L}^{c}_{cls}(\br{x}_c,\br{y}_c)+\mc{L}^{h}_{cls}(\br{x}_h,\br{y}_h),\\
&=-\sum_i \br{y}^c_i \log (\br{p}^c_i) -\sum_j \br{y}^h_j \log (\br{p}^h_j). \\
\end{split}
\end{equation}

\textbf{Unknown identity rejection loss.} The UIR loss $\mc{L}_{uir}$ aims to find a feature reprojection with unsupervised regularization between different domains. Inspired by~\cite{yu2019unknown}, we develop the unknown identity rejection loss $\mc{L}^{c}_{uir}$, for each batch sample $\mc{B}^c$, taking $1/4$ human face data $\br{x}_h$ as unlabeled data and $3/4$ of $\br{x}_c$ as known ones. $\mc{L}^{h}_{uir}$ is defined respectively. These two combined losses can be expressed as:
\begin{equation}
\begin{split}
\mc{L}_{uir}&=\mc{L}^{c}_{uir}(\br{x})+\mc{L}^{h}_{uir}(\br{x}),\\
&=-\sum_{\br{p} \in \mc{B}^c} \log(\frac{\br{p}_i}{\sum_j \br{p}_j})-\sum_{\br{p}' \in \mc{B}^h} \log(\frac{\br{p}'_i}{\sum_j \br{p}'_j}), \\
&\forall (\br{x}_c) \in \mc{D}_c, (\br{x}_h) \in \mc{D}_h. \\
\end{split}
\end{equation}

\textbf{Domain adaption loss.} To reduce the domain gap between $\mathbb{H}$ and $\mathbb{C}$, we adopt the reciprocal of binary softmax loss to constrain the domain correlation between the cartoon and the human face dataset. The final loss function has the form:
\begin{equation}
\mc{L}_{da}=\Sigma_{\mathbb{H}, \mathbb{C}} \frac{1}{-\br{z_i}\log(\br{D}(\br{x_i}))-(1-\br{z_i})\log(\br{D}(\br{x_i}))},
\end{equation}
where $\br{D}(\cdot)$ denotes the network prediction of the domain adaption classifier. In other words, if the smaller the loss, the better generalization of two domains $\mathbb{H}$ and $\mathbb{C}$. The total loss $\mc{L}_{sum}$ function can be formally presented as:
\begin{equation}
\mc{L}_{sum}=\alpha \cdot \mc{L}_{cls}+\beta \cdot \mc{L}_{uir}+\gamma \cdot \mc{L}_{da}.
\end{equation}
In the above expression, $\alpha$, $\beta$, and $\gamma$ are the corresponding proportional weights of the three loss functions, respectively.

\section{Experiments}

\subsection{Experimental Protocol}
\textbf{Datasets}. We adopt the CASIA-WebFace~\cite{yi2014learning} as human face dataset. In this dataset, we randomly selected 168,266 images of 5,000 people as the human domain data. Besides this dataset, we make use of the proposed iCartoonFace of as cartoon training set, consisting of 389,678 images of 5,013 identities. The test set is composed of 20,000 images of 2,000 identities, accompanied with 2,500 distracters.

\textbf{Implementation}. To make a fair benchmarking, we uniformly employ DenseNet-169~\cite{huang2017densely} as our backbone network for all the classifiers,~\ie, SoftMax, SphereFace~\cite{liu2017sphereface}, CosFace~\cite{wang2018cosface}, ArcFace~\cite{deng2019arcface}, Focal loss~\cite{lin2017focal}.
The training images are uniformly resized to $256 \times 256$ in this paper. We first adopt the sotfmax loss the cartoon face dataset as the pretrained model to initialize our training. Based on this pretrained model, we then use the proposed three losses to jointly train on the cartoon and human face datasets. Specially, the cartoon face regions in all experiments are enlarged as 2 times width and 2 times length.
We use stochastic gradient descent (SGD) as the optimization function. The initial learning rate is set as 0.1, and every 20 epochs the learning rate decay as $0.1$ times of the previous epoch. We empirically set $\alpha=1$,$\beta=0.1$,$\gamma=10$ to balance three weights in the same order of magnitudes.

\begin{figure*}[t]
	\includegraphics[width=1.0\linewidth]{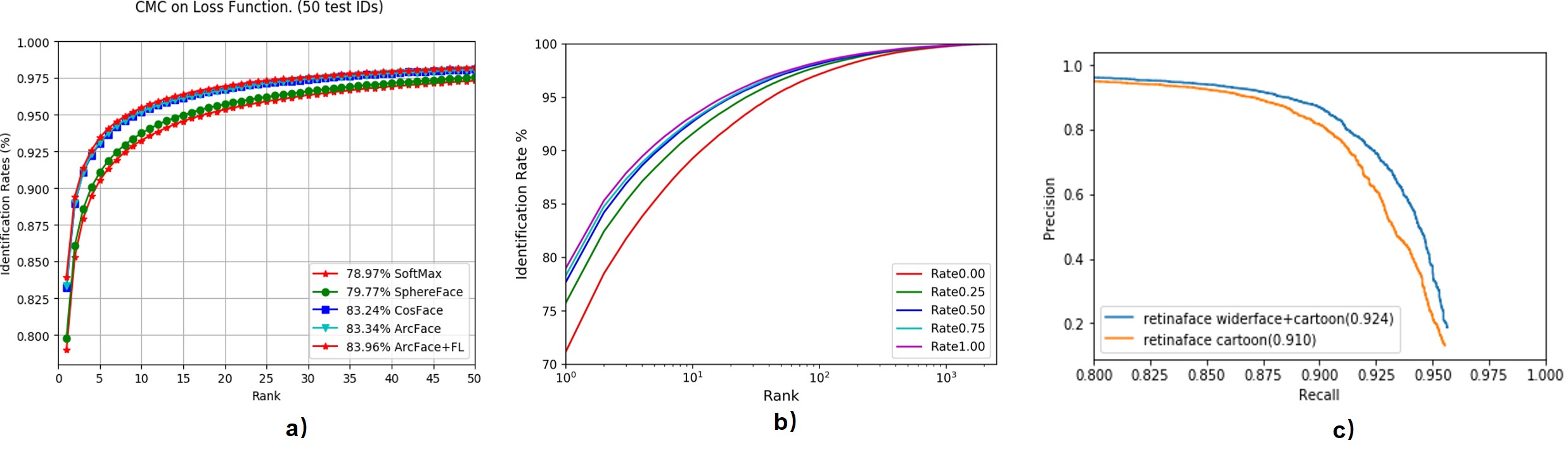}
	\caption{Visualization results. a) CMC curve of five state-of-the-art recognition algorithms. b) CMC curve of recognition with different enlarged rates. c) Precision-Recall on detection dataset for data filtering. }
	\label{fig:clscurve}
\end{figure*}

\begin{table}[t]
	\noindent
	\centering
\renewcommand{\arraystretch}{1.1}
\setlength{\tabcolsep}{0.8mm}
	\caption{Accuray (\%) of state-of-the-art methods on public dataset. Results on human faces are reported by~\cite{wang2020mis}. }
	\label{tab:dataset}
		\begin{tabular}{c|ccc|c}
			 \toprule
            &\multicolumn{3}{c}{Human}&  Cartoon\\
			        &LFW(v)&CFP(v)&  MegaFace &iCartoonFace  \\
            \midrule
            \#images &13k&7k& 1,000k  &389k\\
            \midrule
	        SoftMax &99.59&94.04&93.94&	78.97	  \\
            SphereFace&99.65& 94.22&94.18&	79.77 \\
            CosFace &99.71&95.68&97.69 &83.24	\\
            ArcFace	&99.76&95.28&97.28&83.34	\\
            F-ArcFace&99.71&95.62&97.51 &83.96\\
        \bottomrule
		\end{tabular}

\end{table}

\begin{figure}[t]
	\includegraphics[width=1\linewidth]{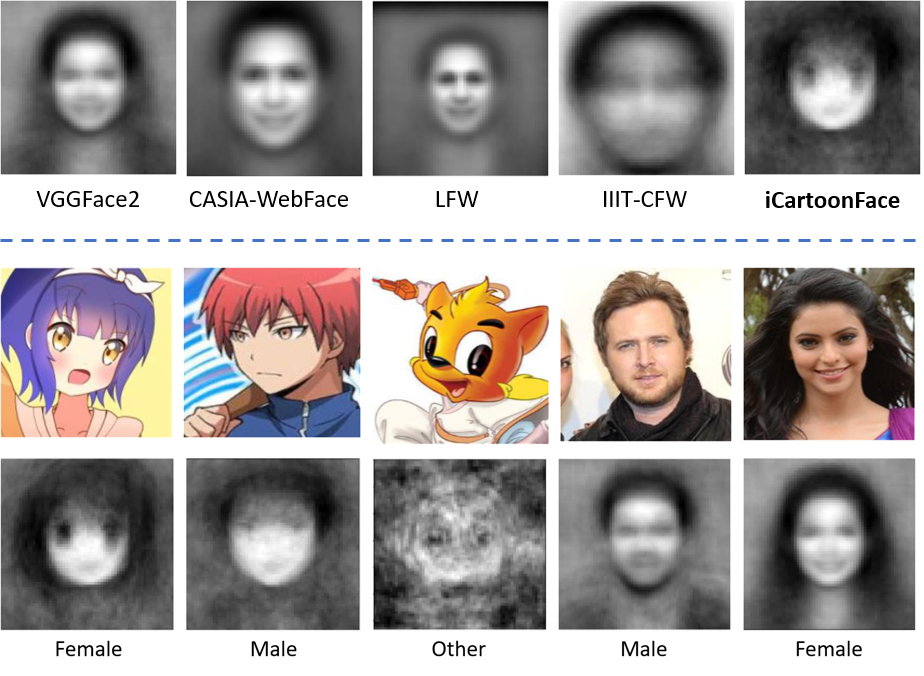}
	\caption{Top: Meanface of three human face dataset and two cartoon dataset. Bottom: representative samples and mean faces in our dataset. The last two human faces are calculated by VGGFace2~\cite{Cao18}. }
	\label{fig:meanface}
\end{figure}

\subsection{Comparisons and Relations}
In~\tabref{tab:dataset}, we exhibit several widely-used face recognition dataset,~\ie, CFP~\cite{sengupta2016frontal}, LFW~\cite{huang2008labeled} and MegaFace~\cite{kemelmacher2016megaface}.
It is notable that LFW and CFP datasets are composed of a limited number of images with 13k and 7k images. This limited scale makes state-of-the-art recognition algorithms~\cite{liu2017sphereface,wang2018cosface,deng2019arcface,lin2017focal} undifferentiated performance. $(v)$ indicates the accuracy on verification set, otherwise identification set.
With the proposal of large-scale dataset~\cite{kemelmacher2016megaface}, various algorithms achieved a rapid development even on million-level dataset, indicating the aligned frontal human faces can be basically solved by the existing techniques. However, in the domain of cartoon face, although notable progress can be achieved by the latest techniques~\cite{deng2019arcface,lin2017focal}, there is still a performance gap when comparing to the human domain. For example, the ArcFace performs over 95\% on all human datasets while only 83.34\% on cartoon faces.

Starting from another perspective, we visualize the mean face of different types of datasets in~\figref{fig:meanface}.
The first three are from the human faces, which are carefully aligned with notable features. For the last two faces of cartoon set, the facial landmarks are thus inconspicuous which brings us new challenges, including the gender issue on the bottom.

\begin{table}[t]
	\noindent
	\centering
\setlength{\tabcolsep}{3.0mm}
	\caption{Accuracy (\%) on state-of-the-art models. The best performance are view in bold.}
	\label{tab:classifier}
		\begin{tabular}{c|ccc}
			 \toprule
			Models &Rank@1 &  Rank@5 & Rank@10\\
            \midrule
	        SoftMax &	78.97	&90.51&	93.24   \\
            SphereFace &	79.77&	91.07&	93.74 \\
            CosFace &	83.24	&93.05	&95.19\\
            ArcFace	&83.34	&93.12	&95.20\\
            F-ArcFace	&\textbf{83.96}	&\textbf{93.43}	&\textbf{95.46}\\
        \bottomrule
		\end{tabular}

\end{table}

\subsection{Performance Analysis}
\textbf{Which is the best algorithm for cartoon face?} To fairly evaluate these algorithms, we integrate the same backbone with five algorithms: softmax, SphereFace~\cite{liu2017sphereface}, CosFace~\cite{wang2018cosface}, ArcFace~\cite{deng2019arcface}, Focal loss~\cite{lin2017focal} with Arcface (F-Arcface). Compared to the basic softmax,~\cite{liu2017sphereface,deng2019arcface,wang2018cosface} improves performance steadily, with over 5\% in rank\@1. In addition, we visualize the CMC curve in~\figref{fig:clscurve}, it can be found that F-Arcface shows a leading performance among the state-of-the-art methods, including the low-rank rate (\eg, rank@1) and higher rate (rank@50). This verifies that the improvements of the algorithm are consistent in all stages for identification.

\begin{table}[t]
	\noindent
	\centering
\renewcommand{\arraystretch}{1.0}
\caption{Ablations of the proposed approach. Our full model reaches the highest performance.}
	\label{tab:ablation}
\setlength{\tabcolsep}{2.6mm}
		\begin{tabular}{ccc|ccc}
			 \toprule
		$\mc{L}_{cls}$&$\mc{L}_{uir}$&$\mc{L}_{da}$ &Rank@1 &  Rank@5 & Rank@10\\
            \midrule
	    \cmark& & &	81.46	&91.94	&94.30   \\
        \cmark&\cmark&    &	81.67&	92.15 &	94.50 \\
         \cmark&\cmark& \cmark  $$ &	\textbf{84.34}&	\textbf{93.52}	&\textbf{96.14}\\

        \bottomrule
		\end{tabular}
	
\end{table}

\textbf{Can human face knowledge transfer to cartoon face?} To evaluate the effectiveness of our different proposed loss functions, we visualize the cartoon recognition task in~\tabref{tab:ablation}. It can be found that the performance of joint cartoon and human face recognition dataset training directly will be worse, compared to only cartoon dataset is used for training in~\tabref{tab:classifier}.  After adding the unknown identity rejection loss and selecting the better hyperparameter $\beta$, the performance will be slightly improved. Finally, a domain transfer loss is added to make the face domain and cartoon domain unified, which can greatly improve the performance of cartoon face recognition.

Despite the recognition task, we also utilize the human domain data for the detection task. We validate the effectiveness of our semi-automatic labeling procedure, which can be found in~\figref{fig:clscurve} c). The first trained cartoon face model reaches an accuracy of 91.0\%. After including the widerface dataset of human beings, the final performance reaches 92.4\%, which helps the annotation process.

\begin{table}[t]
	\noindent
	\centering
\renewcommand{\arraystretch}{1.0}
\setlength{\tabcolsep}{3.5mm}
	\caption{Recognition performance with different enlarged rates of facial region.}
	\label{tab:rate}
		\begin{tabular}{c|ccc}
			 \toprule
			Enlarged Rate &Rank@1 &  Rank@5 & Rank@10\\
            \midrule
	        0.00  & 71.15&	85.26&	89.26   \\
            0.25 &	75.73	&88.30	&91.58\\
            0.50	&77.67	&89.72	&92.69 \\
            0.75 &	78.29	&89.95&	92.86 \\
            1.00	& \textbf{78.97}	&\textbf{90.51}	& \textbf{93.24}\\
        \bottomrule
		\end{tabular}

\end{table}

\textbf{Is context useful for cartoon face recognition?} Cartoon face is the main discriminative part in identifying one character.
However, separating from human faces based on cartoon faces may not be enough to distinguish different cartoon persons in some cases. To explore this, we compare the effects of different expansion proportions by $[0.0,1.0]$ to introduce more context information, such as haircut and decorations. In this setting, we use the same DenseNet-169 as the backbone network and softmax losses. From~\tabref{tab:rate}, it can be easily obtained that more contextual information can be beneficial to the identification task. This verifies that in the cartoon person recognition task, the face region plays the most important role in identifying a cartoon person. And the CMC curve in~\figref{fig:clscurve} b) also indicates the higher contextual information is included, the higher performance can be achieved.

\section{Conclusion}
In this paper, we take a closer look into the cartoon media by establishing a benchmark dataset, namely iCartoonFace. The dataset shows many meaningful features, including high-quality, large-scale, in-the-wild, and with rich annotations. Beyond these issues, we thus designed to investigate the challenges and present two typical tasks,~\ie, face recognition, and face detection. Under this setting, we present a multi-task domain adaptation solution to utilize the human data to cartoon media. Experimental evidences verify the potential of multi-domain exploitation and analyze the data benchmark. We believe that the research on cartoon face recognition would bring more attractive researches and broad industrial applications.
\bibliographystyle{ACM-Reference-Format}
\bibliography{ref}
\end{document}